\documentclass[times, 10pt,twocolumn]{article}
\usepackage{latex12}
\usepackage{times}
\usepackage{balance}
\usepackage{graphicx}

\begin{document}

\title{Semantic Segmentation of highly class imbalanced fully labelled 3D volumetric biomedical images and unsupervised Domain Adaptation of the pre-trained Segmentation Network to segment another fully unlabelled Biomedical 3D
Image stack}

\author{1. Shreya Roy 2. Prof. Anirban Chakraborty\\
\emph{MTech(Research) student at Indian Institute of Science, Bangalore(1)}\\  \emph{ Assistant Professor at Indian Institute of Science, Bangalore(2)}\\
\emph{shreyaroy@iisc.ac.in}\\
% For a paper whose authors are all at the same institution,
% omit the following lines up until the closing ``}''.
% Additional authors and addresses can be added with ``\and'',
% just like the second author.
%\and
%Second Author\\
%Institution2\\
%First line of institution2 address\\ Second line of institution2 address\\
%SecondAuthor@institution2.com\\
}

\maketitle
\thispagestyle{empty}

\begin{abstract}
The goal of our work is to perform pixel label semantic seg-
mentation on 3D biomedical volumetric data. Manual an-
notation is always difficult for a large bio-medical dataset.
So, we consider two cases where one dataset is fully labeled
and the other dataset is assumed to be fully unlabelled. We
first perform Semantic Segmentation on the fully labeled
isotropic biomedical source data (FIBSEM) and try to incor-
porate the the trained model for segmenting the target unla-
belled dataset(SNEMI3D)which shares some similarities with
the source dataset in the context of different types of cellular
bodies and other cellular components. Although, the cellular
components vary in size and shape. So in this paper, we have
proposed a novel approach in the context of unsupervised do-
main adaptation while classifying each pixel of the target vol-
umetric data into cell boundary and cell body. Also, we have
proposed a novel approach to giving non-uniform weights to
different pixels in the training images while performing the
pixel-level semantic segmentation in the presence of the cor-
responding pixel-wise label map along with the training orig-
inal images in the source domain. We have used the Entropy
Map or a Distance Transform matrix retrieved from the given
ground truth label map which has helped to overcome the
class imbalance problem in the medical image data where the
cell boundaries are extremely thin and hence, extremely prone
to be misclassified as non-boundary.

%\keywords{Unsupervised Domain Adaptation, Semantic Segmentation for highly class imbalanced medical image}
\end{abstract}
\section{Introduction}
\label{sec:intro}
In the bio-medical domain, the fully annotated images are not always available. The UNeT developed by Olaf Ronneberger et al. ~\cite{ronneberger2015u} for Biomedical Image Segmentation became very popular as it relies on the strong use of data augmentation to use the available annotated samples more efficiently.It consists of two paths - contraction path (also called as the encoder) which captures the context in the image and the second path the symmetric expanding path (also called as the decoder)which localizes the context in the image. UNet can accept the image of any size as it has no dense layer.\par In the UNet, we have skip connection between layers of Encoder and Decoder of the same level so that the small object information which might get vanished during downsampling can be added as a residual during the Upsample phase.\par
Also, Bio-Medical Image segmentation involves the volume of 3D cross-sectional slices where each pixel on every slice is to be labeled as a class object. There are problems of anisotropy involved with the 3D image stack as the resolution of the ion beam varies with depth thus producing cross-section images with varying intensity and resolution. Hence, accurate segmentation of 3D EM images requires not only accurate segmentation of 2D slices but also making consistent predictions across slices to deal with
misalignment problems.To perform semantic segmentation on 3D volumetric biomedical image data, a 3D architecture was proposed in 2016 in the paper 3D U-Net: Learning Dense Volumetric Segmentation from Sparse Annotation\cite{cciccek20163d}. But, the major drawback with 3D architecture for semantic segmentation is computational expensiveness.In the paper 'DeepEM3D: approaching human-level performance on 3D anisotropic EM image segmentation' \cite{zeng2017deepem3d}the authors proposed to employ a combination of novel boundary map generation methods with optimized model ensembles (2D-3D Ensembling) to address the challenges of segmenting anisotropic images. They trained multiple deep models variants that accepted different numbers of input slices and predicted boundaries of different thickness suppressing noise in Z-direction alignment. Due to the limitations of both GPU memory and computing power, when designing 2D/3D CNNs for 3D biomedical image segmentation, the trade-off between the field of view(achieved by 2D CNN) and utilization of inter-slice information in 3D images(achieved by 3D CNN) remains a major concern. Also, their method requires a little human labor to identify the misalighen slices.
\par This limitation was addressed in the paper 'Combining Fully Convolutional and RecurrentNeural Networks for 3D Biomedical Image Segmentation'\cite{chen2016combining}where the authors proposed BiDirectional LSTM on the top of the FCN to capture the inter slice information. They used a different variety of UNet called as k-UNet to extract the intra-slice information at different scales sequentially(from the coarsest scale to the finest scale). The information extracted by the submodule FCN responsible for a coarser scale will be propagated to the subsequent submodule FCN to assist the feature extraction in a finer scale. In our experiment, we have used the simple 2D UNet architecture to perform semantic segmentation on each slice independently at the expense of interslice context information as our main objective is to make the UNet adapt to another domain in an unsupervised fashion. We wanted to avoid further complexity of training the complex UNet model in the unsupervised fashion after the pretraining phase. 
\begin{figure}
\label{sec:fig}
\begin{minipage}[b]{1\linewidth}
  \centering
  \centerline{\includegraphics[width=8.5cm]{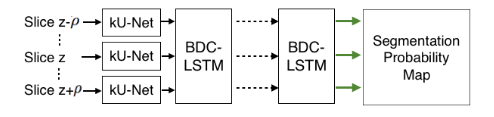}}
%  \vspace{1.5cm}
  \centerline{ Fig 0:BDC-LSTM model to capture inter-slice context }\centerline{information \cite{chen2016combining}}\medskip
  \end{minipage}
  \end{figure}
 \ref{sec:fig}
 \par
Also, Bio-Medical Images are highly class imbalanced. In our case, the main objective is to segment the image into the cell boundary and cell body. Cell boundary is so thin that during classification the UNet is very likely to label them as cell bodies. This challenge we have overcome in a novel manner (details are in the Methodology section). Also, we have performed unsupervised domain adaptation on our segmentation network where we considered the target domain data to be fully unlabeled.\par In a brief, we trained a 2D UNet on the slices of FIBSEM data, and then, we try to perform unsupervised domain transform on the target data SNEMI3D.
\begin{figure}
\label{sec:fig}
\begin{minipage}[b]{1\linewidth}
  \centering
  \centerline{\includegraphics[width=8.5cm]{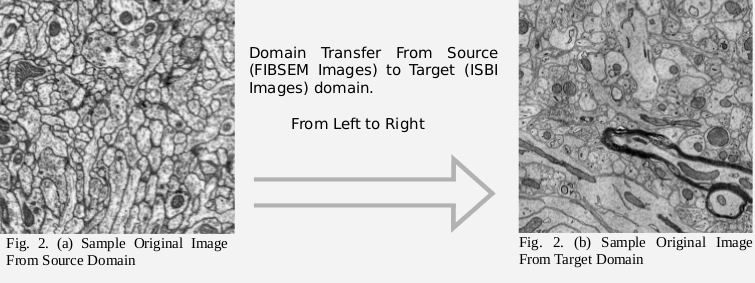}}
%  \vspace{1.5cm}
  \centerline{ Fig 1:(a) Figure illustrates the domain adaptation from }\centerline{Source to Target}\medskip
  \end{minipage}
  \end{figure}
 \ref{sec:fig}
\begin{figure}[htb]
\label{sec:fig300}
\begin{minipage}[b]{1\linewidth}
  \centering
  \centerline{\includegraphics[width=8.5cm]{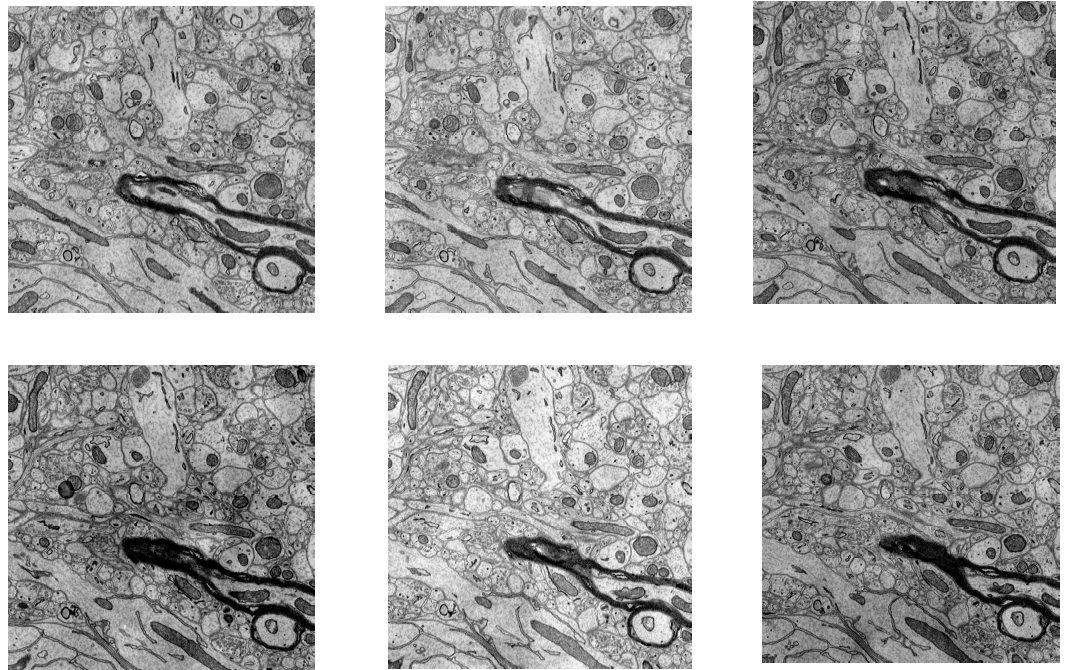}}
%  \vspace{2.0cm}
  \centerline{Fig 1:(b) SNEMI3D data (Anaisotropic): The resolution and } \centerline{intensities vary among slices}\medskip
\end{minipage}
\end{figure}
\ref{sec:fig300}
\section{Related Work}
In recent days, there have been a lot of studies of domain adaptation for biomedical images. In the paper \cite{bermudez2018domain}A domain-adaptive two-stream U-Net for electron microscopy image segmentation the authors proposed a two stream UNet, one for the source domain and one for the target, and allowing some of their weights to differ, while the others are shared. The main idea was a compromise between preserving what can be learned from the source domain using enough training data and adapting the weights to the potentially different statistics of the target domain. A slight modification of this approach was proposed in \cite{roels2019domain}Domain Aaptive Segmentation In Volume Electron Microscopy Imaging where a second decoder is attached to the encoder-decoder segmentation network which reconstructs both the source and target data. Although, our approach to handle the domain adaptation is different from the ones mentioned above. We try to learn the label distribution of the source dataset in an adversarial manner as the final segmentation result is boundary and non boundary for both the domain and hence, we try to minimize the gap between the label distribution between source and target domain.
\begin{figure}
\label{sec:fig4}
\begin{minipage}[b]{1\linewidth}
  \centering
  \centerline{\includegraphics[width=8.5cm]{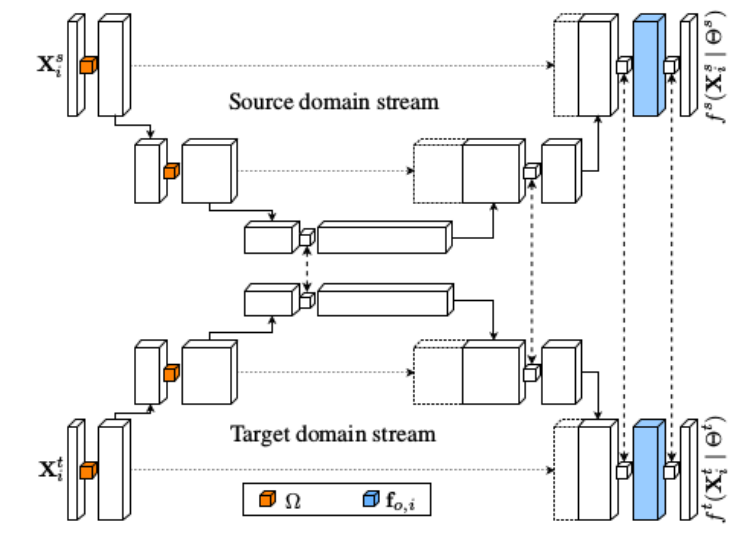}}
%  \vspace{1.5cm}
  \centerline{Fig 2: Two Stream UNet architecture\cite{bermudez2018domain}}\medskip
  \end{minipage}
  \end{figure}
 \ref{sec:fig4}
\section{Dataset Description}In the
FIBSEM dataset, the images are generated from a 5 microm-
eter cube volume of D. Melanogaster larval neuropil imaged
at 10x10x10nm resolution using focused ion beam scanning
electron microscopy. The dataset contains 500 cross-section
slices of the 3D volume, where each slice contains 500X500
pixels. Pixel Label annotation map was provided for this
dataset where each pixel in the volume is identified as cell
boundary (1), cytoplasm (2), mitochondria (3), or glial cell
(4). The resolution and intensities do not vary among slices.
Then we try to perform unsupervised domain transform on
the target data SNEMI3D, a 6X6X30 nm per voxel resolution serial section scanning EM (ssSEM) volume which covers a
micro-cube of approximately 6 X 6 X 3 microns.

\section{Methodology}
\subsection{Formulation of Entropy Map for each pixel from given training labelmaps(Experiment1)-} First, we had to deal with the class imbalance problem while performing
segmentation on the FIBSEM data as the cell boundary contains almost 10 percent of total pixels in a
2D slice. Here, we introduce a novel approach of using entropy information of a pixel as weights in the
classification layer of the UNet. The entropy of a pixel is calculated computing the entropy of a 5X5 kernel
matrix centered on the target pixel, the elements of which are the class labels of the neighboring pixels.
The idea is to give more weights to the pixels which have more entropy (i.e., different labels are assigned
in its neighboring pixels) while classifying a pixel as one of the 4 classes in the ground truth label map.
\begin{figure}[htb]
\label{sec:fig0}
%\begin{minipage}[b]{1.0\linewidth}
  \centering
  \centerline{\includegraphics[width=8.5cm]{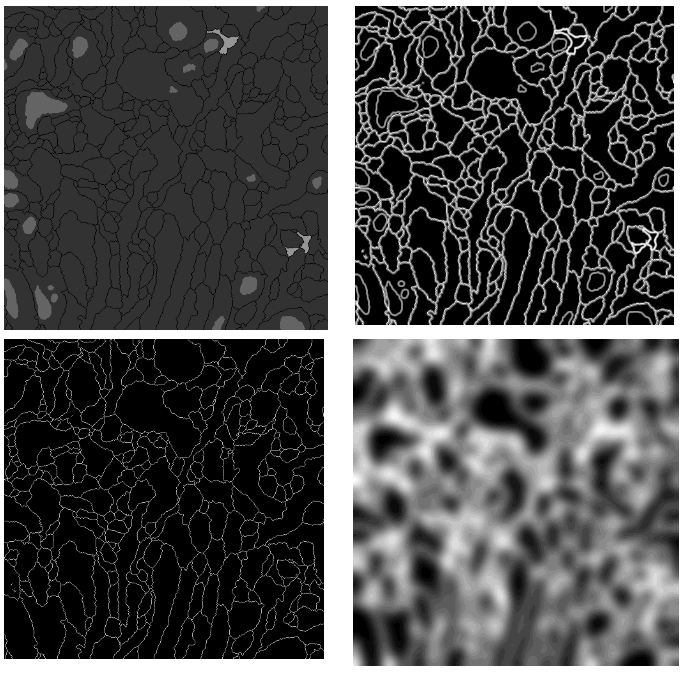}}
%  \vspace{2.0cm}
  \centerline{Fig 3- Ground Truth Label Map Entropy}\centerline{  Weight Map,Distance Tranform matrix}\centerline{Smoothed Distance Transform Matrix}\medskip
%\end{minipage}
\end{figure}
\ref{sec:fig0}
\subsection{Formulation of Distance Transform matrix from training label maps(Experiment2)-}As the main task is to find cell boundary and non-boundary in the target dataset which is a binary classification problem, we first performed a similar binary classification on
FIBSEM data (cell boundary and the rest 3 classes as non-boundary). For this binary classification task, we had a similar problem of class imbalance which was taken care of by using the Distance Transform of the original boundary map. Also, we had to smooth the distance transform
matrix with a Gaussian kernel of sigma=10 in order to avoid giving 0 weights to most of the nonpboundary pixels.

\subsection{Training the pretrained UNet(On labeled source domain) in adversarial fashion(Experiment3)-}Here, we use the trained model from Experiment 2 as the initial segmentation
model and then try to learn the label distribution of boundary and non-boundary in the source domain in
an adversarial fashion as we assume that the images in the target domain follow almost a similar boundary non-boundary distribution. To achieve the goal we need a discriminator that will identify the true source
label map as true and generated label map from the segmentation UNet as false for the target domain.\par So, we have 3 loss components: Supervised loss on Source domain (say L1) which was used to train the segmentation network UNet in Experiment 2 for boundary non-boundary prediction, and the losses incurred by the Discriminator while being trained on the source ground truth as true (say L2) and UNet generated label map of the images in target domain as false (say L3) respectively. Now, we will train the UNet to maximize the Discriminator's loss L3. We will no longer use the Supervised Loss L1 to update the parameters of the UNet as it is trying to find the embedding for the target distribution only.
\par
Basically, we are using the weights of trained UNet (trained on L1) to initialize the weights for our
segmentation network for the target domain and learn the weights of the network for the target domain
in an adversarial manner. As the adversarial training progresses, the network will capture the Target
domain’s embedding.
\begin{figure}[htb]
\label{sec:fig1}
%\begin{minipage}[b]{1\linewidth}
  \centering
  \centerline{\includegraphics[width=8.5cm]{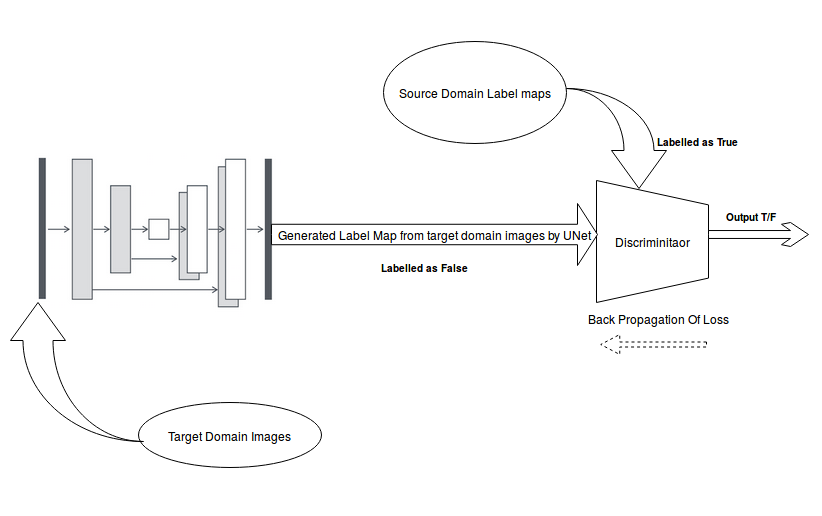}}
%  \vspace{1.5cm}
  \centerline{Fig 4:(a) Discriminator Training}\medskip
%\end{minipage}
\end{figure}
\ref{sec:fig1}
\begin{figure}[htb]
\label{sec:fig2}
%\begin{minipage}[b]{1\linewidth}
  \centering
  \centerline{\includegraphics[width=8.5cm]{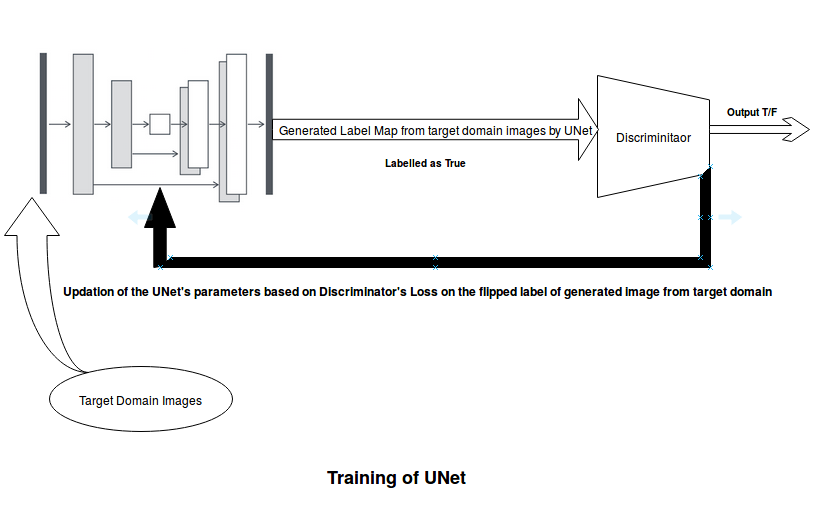}}
%  \vspace{1.5cm}
  \centerline{Fig 4:(b) Training of Pretrained UNet on Discriminator's loss}\centerline{in an adversarial fashion}\medskip
%\end{minipage}
\end{figure}
\ref{sec:fig2}
\section{ Result-}
\subsection{ Experiment 1-} After 120 iterations with initial learning rate of .0001 in the first experiment (Experiment 1) on FIBSEM, we receive the model M1 which achieves the validation Jaccard Score 72 \% which is quite satisfactory considering multi-class classification of every pixel in the
medical images where cell boundary class is extremely thin and dominated by the other 3 classes. As our primary
goal was to find cell boundary and non-boundary in the target domain (SNEMI3D), we did not continue this supervised segmentation task with 4 classes further.
\subsection{Experiment 2-} To obtain the model for determining the only boundary and non-boundary (Say
M2) on FIBSEM we initialize the weights of all layers (excepting the classification layer)of the 2D
UNet with the trained model obtained from experiment 1, i.e., M1 and after 20 iterations the model M2
is achieved with similar Jaccard Score of 80 \% which is quite good considering we used only 2D UNet
for 3D volume segmentation which can not capture the interslice information. We did not go for 3D
UNet as it will be computationally expensive and our primary goal is not to perform semantically
segmentation on Source Domain Images. In this setup, we used the Distance transform matrix as the
pixel weight map to find the weighted cross-entropy.\par
Even, if we only train to predict boundary and nonboundary on FIBSEM without using the model
M1, after 90 iterations we obtain a similar accuracy.\par
Another observation is that to tackle the problem of class imbalance if we had used weights for
boundary and nonboundary with specific ratios (say 2:1, 5:1 or 10:1), after the same number of
iterations none of them achieve as good as the Jaccard Score achieved with the distance transform
the matrix used as a weight map. As target domain has varying intensity and resolution across the depth of the
slices, we train the model M2 for another 10 iterations after applying some random Jitter transform on
the input image slices and the obtained model M3 works as an initial segmentation model for target data.

\begin{figure}[htb]
\label{sec:fig202}
  \centering
  \centerline{\includegraphics[width=8.5cm]{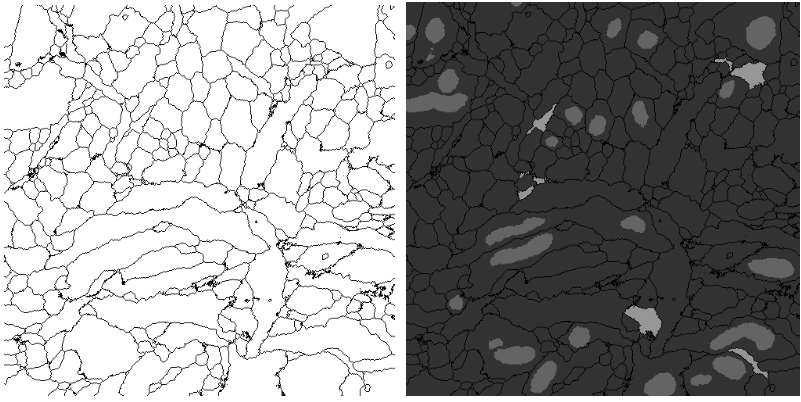}}
%  \vspace{2.0cm}
  \centerline{Fig 5: (a) Original 4-class Label map and 2 class Boundary map}\medskip

\end{figure}
\ref{sec:fig202}
\begin{figure}[htb]
\label{sec:fig200}
  \centering
  \centerline{\includegraphics[width=8.5cm]{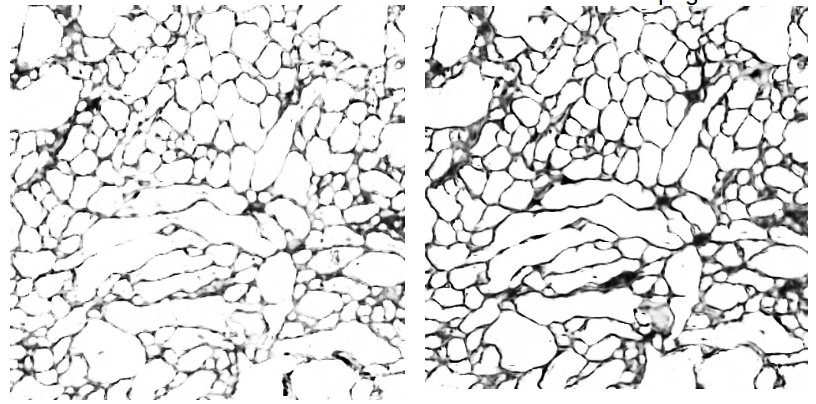}}
%  \vspace{2.0cm}
  \centerline{Fig 5:(b) Result of Experiment 2- 2:1 and  5:1}
  \medskip

\end{figure}
\ref{sec:fig200}
\begin{figure}[htb]
\label{sec:fig201}
  \centering
  \centerline{\includegraphics[width=8.5cm]{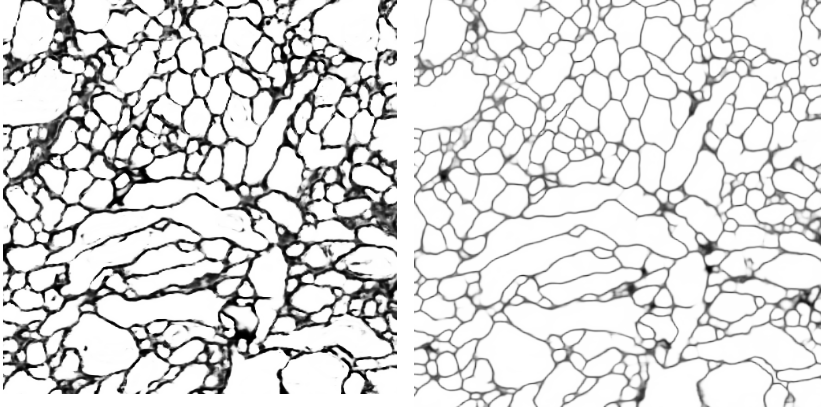}}
%  \vspace{2.0cm}
  \centerline{Fig 5:(c) Result of Experiment 2- using 10:1 }\centerline{smoothed distance transformed matrix}\medskip

\end{figure}
\ref{sec:fig201}

\subsection{Experimen 3-}
Results after each iteration are shown in the table for an image sampled from
the target domain. Here, we did not use the label information of target domain images to train the
network as our main purpose is to make the network M3 obtained from Experiment 2 adapt to the target
domain in an unsupervised manner assuming no label information was provided for the images in the
target domain.

\begin{figure}[htb]
\label{sec:fig900}
%\begin{minipage}[b]{1\linewidth}
  \centering
  \centerline{\includegraphics[width=8.5cm]{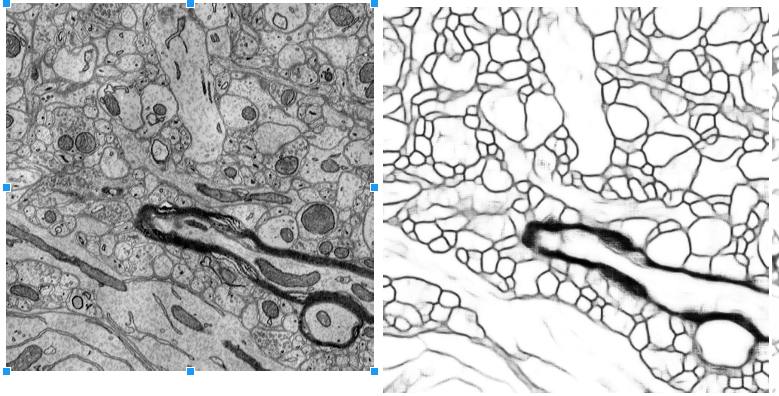}}
%  \vspace{2.0cm}
  \centerline{Fig 6:(a) Result of Experiment 3 Original Image sample and }\centerline{ corresponding Original Label Map from target domain}\medskip
%\end{minipage}
\end{figure}
\ref{sec:fig900}
\begin{figure}[htb]
\label{sec:fig901}
\begin{minipage}[b]{1\linewidth}
  \centering
  \centerline{\includegraphics[width=8.5cm]{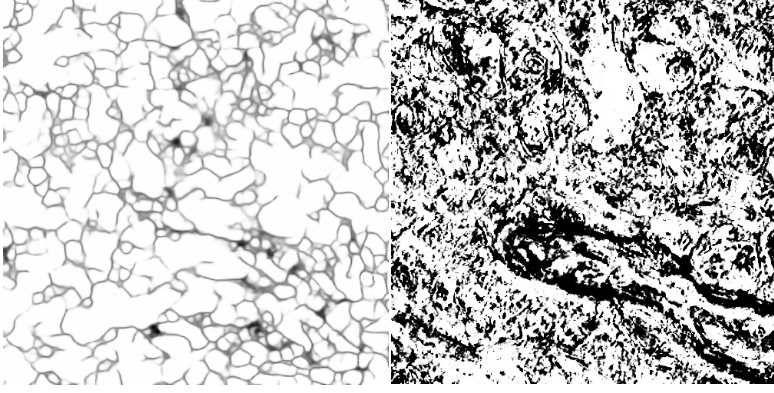}}
%  \vspace{2.0cm}
  \centerline{Fig 6:(b) Output Image after 1st and 2nd epoch of }\centerline{ adversarial training}\medskip
\end{minipage}
\end{figure}
\ref{sec:fig901}
\begin{figure}[htb]
\label{sec:fig902}
\begin{minipage}[b]{1\linewidth}
  \centering
  \centerline{\includegraphics[width=8.5cm]{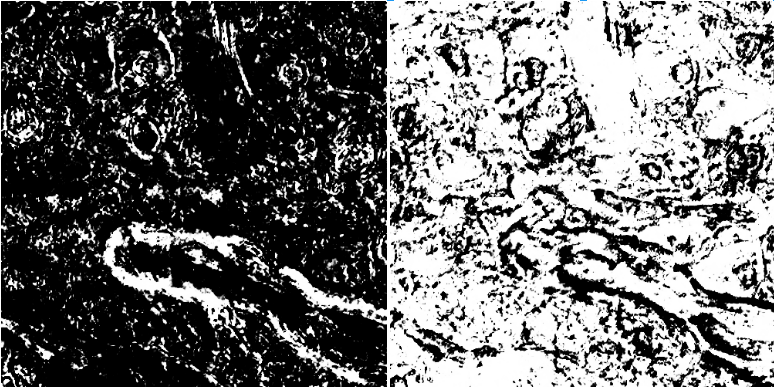}}
%  \vspace{2.0cm}
  \centerline{Fig 6:(c) Output Image after 3rd and 4th epoch of}\centerline{  adversarial training}\medskip
\end{minipage}
\end{figure}
\ref{sec:fig902}
\section{Future Work}
We are planning to impose condition on the Discriminator where we can give a pair of inputs: Input Image and corresponding true Label Map for the source domain as true (say loss contributed by these set of examples as L1), Input Image and Label Map of another slice (negative example) for the source domain as False (say loss contributed by these set of examples as L2), and Input Image and corresponding generated Label Map for the target domain as false (say loss contributed by these set of examples as L3) and the segmentation network will only be trained on negative of L3 ie., the Input Image and corresponding generated Label map as True. In this way, we can learn the joint embedding of both the Input Image and corresponding label map. 
\par
Also, we plan to learn the joint embedding of both the source and target domain as our present experiment is moving the network away from the source domain’s embedding, it’s not finding out the common embedding between source and target data.

%-------------------------------------------------------------------------

\bibliography{latex12.blb}{}
\bibliographystyle{latex12}
\end{document}